\documentclass{article}
\usepackage{kr}
\usepackage{times}
\usepackage{soul}
\usepackage{url}
\usepackage[hidelinks]{hyperref}
\usepackage[utf8]{inputenc}
\usepackage[small]{caption}
\usepackage{graphicx}
\usepackage{amsmath, amssymb}
\usepackage{mathtools}
\usepackage{amsthm}
\usepackage{booktabs}
\usepackage{algorithm}
\usepackage{algorithmic}
\usepackage{array}

\newtheorem{example}{Example}
\newtheorem{definition}{Definition}

\title{Applying Relation Extraction and Graph Matching\\to Answering Multiple Choice Questions}

\author{
Naoki Shimoda$^1$\and
Akihiro Yamamoto$^{1,2}$\\
\affiliations
$^1$Graduate School of Informatics, Kyoto University\\
$^2$Center for Innovative Research and Education in Data Science,\\Kyoto University\\
\emails
\{shimoda.naoki.77s@st, akihiro@i\}.kyoto-u.ac.jp
}

\begin{document}

\maketitle


\begin{abstract}
    In this research, we combine Transformer-based relation extraction with matching of knowledge graphs (KGs) and apply them to answering multiple-choice questions (MCQs) while maintaining the traceability of the output process.
    KGs are structured representations of factual knowledge consisting of entities and relations.
    Due to the high construction cost, they had been regarded as static databases with validated links.
    However, the recent development of Transformer-based relation extraction (RE) methods has enabled us to generate KGs dynamically by giving them natural language texts,
    and thereby opened the possibility for representing the meaning of the input sentences with the created KGs.
    Using this effect, we propose a method that answers MCQs in the ``fill-in-the-blank" format,
    taking care of the point that RE methods generate KGs that represent false information if provided with factually incorrect texts.
    We measure the truthfulness of each question sentence by (i) converting the sentence into a relational graph using an RE method and (ii) verifying it against factually correct KGs under the closed-world assumption.
    The experimental results demonstrate that our method correctly answers up to around 70\% of the questions, while providing traceability of the procedure.
    We also highlight that the question category has a vast influence on the accuracy.
\end{abstract}

\section{Introduction}
In this research, we combine \textbf{relation extraction (RE)} based on the Transformer and matching of \textbf{knowledge graphs (KGs)}, and apply the combination to answering multiple-choice questions where
RE means to extract the relations of entities from given natural language texts.
Since the construction of KGs required massive human annotations, KGs have long been regarded as static databases with only factually correct links.
However, the recent development of Transformer-based RE methods has enabled us to construct KGs dynamically for any text in a natural language, e.g., English, at a low cost.
KGs have a graph structure, and algorithms for graphs are applicable to them.
In particular, deciding whether a KG matches another larger one as a subgraph semantically means the latter KG entails the former.
Therefore, we conjecture that entailment of one sentence in natural language by another could be checked by applying RE to the sentences and graph matching to the obtained KGs.
Moreover, the matching of KGs makes the entailment relation between the sentences traceable.

\textbf{Multiple-choice questions (MCQs)} are widely used in higher education due to their objective evaluation and the ability to conduct exams efficiently on a large scale.
For learners who answered an MCQ test,  receiving the reasons why each option is correct or incorrect after the test is effective for both enhancing the retention of the topic and reducing the misunderstanding of facts caused by reading the false options \cite{butler_feedback_2008}.
However, it is not easy for teachers to create explanations for all the answers. Large language models (LLMs) are expected to reduce their burden by being asked to generate step-by-step explanations leading to the answers.
Unfortunately, in the current status, it is known that such explanations often lead to reasoning errors depending on the properties of the options in MCQs and the order of them \cite{turpin_language_2023}.
Moreover, even though becoming less often, LLMs might make hallucinations by stating non-existent facts, and reasoning based on them makes reliability a persistent issue.\par

Our study tries to overcome these problems by combining a Transformer-based RE method and matching of KGs, and proposes a method for answering MCQs with reliable explanations.
The RE method creates a tailored relational graph for representing the meaning of the question texts of MCQs.
By measuring the truthfulness of the obtained graph and verifying it against factually correct KGs under the closed-world assumption, our method outputs the most reliable word from the given choices with its explanation.\par

This paper is organized as follows.
First, in Section \ref{sec:task-def}, we provide the formal definition of the task that we addressed in this study.
Section \ref{sec:prep} introduces the key concepts and methods that underpin our discussion.
In Section \ref{sec-method}, we describe the details of the proposed method, followed by Section \ref{sec:experiments}, where we present the results of answering experiments conducted on original MCQ datasets.
Section \ref{sec:ex-research} reviews related works in the fields of MCQ answering and fact verification.
Finally, Section \ref{sec-conclusion} concludes the paper and outlines directions for future work.

\section{Multiple Choice Question Answering}\label{sec:task-def}
\textbf{Multiple-choice question answering (MCQA)} is a task where a model is applied to select the correct option given a question sentence $q$, a set of options $\mathcal{O}=\{o_1, \dots, o_k\}$, and a set of background knowledge $\mathcal{C}=\{C_1, \dots, C_k\}$, where $C_i$ is related to each choice $o_i\quad(i=1,\dots,k)$ \cite{shah_what_2020}.
We treat the case of the 4-way cloze test format, where the task is to choose the best word that fits in the blank from four options.

\begin{example}[Barack Obama]\label{ex-barack-obama}
    Let $q(x)$ be \textit{``$\{x\}$ is an American politician and attorney who served as the 44th president of the United States from 2009 to 2017,"} where
    \[
        \mathcal{O} = \left\{\begin{array}{ll}
            o_1  \colon\text{``Barack Obama"},   & o_2  \colon\text{``Joe Biden"},   \\
            o_3  \colon\text{``George W. Bush"}, & o_4  \colon\text{``Bill Clinton"}
        \end{array}\right\}.
    \]
    The correct answer is $o_1\colon\text{``Barack Obama"}$.
\end{example}

\begin{definition}[Cloze test 4-way MCQA]
    \textbf{Cloze test 4-way MCQA} is a task of selecting the correct option $o_i$ given a question sentence $q(x)$ with a blanck $x$, the set of four option words $\mathcal{O}=\{o_1, o_2, o_3, o_4\}$, and a set of background knowledge $\mathcal{C}=\{C_1,C_2,C_3,C_4\}$.
    The task is to find the correct answer $o_i\in\mathcal{O}$ such that
    \begin{equation}
        C_i \models q(o_i),\label{eq-mcqa}
    \end{equation}
    where $\models$ informally indicates the entailment between natural language sentences.
\end{definition}

We require a transformation $\mathcal{G}$ with which
$q(x)$ and $C_i$ are transformed into some objects ${\cal G}(q(x))$ and ${\cal G}(C_i)$, respectively,
and to provide a \textbf{proof method} for deciding
\begin{equation*}
    {\cal G}(C_i) \vdash {\cal G}(q(o_i)),
\end{equation*}
to solve the entailment problem (\ref{eq-mcqa}).
We adopt the set of knowledge graphs (KGs) as the co-domain of $\mathcal{G}$ because a KG is regarded as a set of conjunctions of binary ground atoms in first-order logic.
Then the method of $\vdash$ is the inclusion checking of two conjunctions, which can be seen as the subgraph isomorphism problem in graph terminology.

\section{Preliminaries}\label{sec:prep}
In this section, we introduce the definition of KGs and the various concepts used in the proposed method.

A \textbf{Knowledge Graph (KG)} is a data structure that represents objects such as people, things, dates, and places as nodes and their relations as labeled directed edges \cite{book-KG}.
Focusing on the fact that a knowledge graph $G$ consists of triplets $(v_1, r, v_2)$, where $r\in R$ represents the directed edge from node $v_1$ to $v_2$, we represent $G$ as a set of triplets $\left\{(v_1, r, v_2)\right\}$.

\textbf{Relation Extraction (RE)} is the task of extracting relations between entities described in a natural language sentence, and it is an important step to automatically construct KGs \cite{zhao_comprehensive_2024}.
\begin{example}[Relation Extraction]
    Assume that a sentence \textit{``Barack Obama, who served as the 44th President of the US, was born in Hawaii"} is given, the RE is to output a set of semantic triplets, such as,
    \begin{equation*}
        \left\{
        \begin{aligned}
             & (\mathrm{``Barack\ Obama"}, \mathrm{``born\ in"}, \mathrm{``Hawaii"}),             \\
             & (\mathrm{``Barack\ Obama"}, \mathrm{``is\ a"}, \mathrm{``President\ of\ the\ US"})
        \end{aligned}
        \right\}.
    \end{equation*}
\end{example}
\noindent
In this study, we consider two kinds of RE methods that are reported to achieve the best F1 scores in the NYT dataset \cite{hutchison_modeling_2010,zhao_comprehensive_2024},
namely \textbf{REBEL} \cite{huguet_cabot_rebel_2021} and \textbf{UniRel} \cite{tang_unirel_2022}.

\textbf{Entity Linking (EL)} is the task of mapping a textual mention to its corresponding entry in a structured knowledge base \cite{shen_entity_2015}.
Within the field of knowledge graphs, EL is widely used for \textbf{entity resolution}, i.e., merging nodes that represent the same real‑world entity either within a single KG or across multiple KGs \cite{sun_bootstrapping_2018}.

\textbf{Semantic Textual Similarity (STS)} is a measure of semantic equivalence between the given two blocks of text \cite{chandrasekaran_evolution_2021}.
In this study, we utilize the STS of different node labels as the \textbf{similarity of nodes} in KGs.

\section{Methodology}\label{sec-method}
Our proposing method takes a question sentence $q(x)$ and a set of option words $\mathcal{O}$ as input and estimates the correct option $\hat{o}\in \mathcal{O}$ through the following steps:
\begin{enumerate}
    \setlength{\leftskip}{0.2cm}
    \item Create a \textbf{propositional graph} $\mathrm{PG}(o_i)$ for each choice $o_i$ to represent the relations extracted from the sentence $q(o_i)$.
          We also create a \textbf{knowledge graph} $\mathrm{KG}(o_i)$ for each choice $o_i$ to represent the background knowledge for solving the question.
    \item Estimate the corresponding nodes between $\mathrm{PG}(o_i)$ and $\mathrm{KG}(o_i)$.
    \item Verify each triplet in $\mathrm{PG}(o_i)$ against the ones from $\mathrm{KG}(o_i)$ to compute the \textbf{truthfulness} of each original sentence $q(o_i)\ (i=1,2,3,4)$.
    \item Choose the correct answer based on the truthfulness and the similarity of node labels between $\mathrm{PG}(o_i)$ and $\mathrm{KG}(o_i)$.
\end{enumerate}

Through Steps 1 to 3, fact verification using KG matching is applied to each proposition $q(o_i)$, and the truthfulness of the proposition being true is computed by predefined rules.
The final result is selected in Step 4.
We assume that an RE method $\mathcal{G}$, an entity linking method $\mathcal{L}$, and a semantic similarity measure $\mathrm{sim}(v_1,v_2)$ are pre-defined.
In the following, we describe the details of each step.

\paragraph{Step 1.}~\\
\noindent{\bf Constructing Propositional Graph}
In the first step, we create a relational graph based on each proposition $q(o_i)$ to represent the relations to be verified.
At this time, if $o_i$ is an incorrect option, the relations represented in $q(o_i)$ will also contain factual errors.
Therefore, we call the relational graph created from $q(o_i)$, which could contain incorrect relations, a \textbf{propositional graph} (PG), and distinguish it from the \textbf{knowledge graph} (KG) that only contains the ground truth triplets.
By using an RE method $\mathcal{G}$, we construct such a propositional graph for each choice $c_i\ (i=1,2,3,4)$ as $\mathcal{G}(q(o_i))$.

In the propositional graph for choice $o_i$, it is desirable that the structure of graphs other than the option label $o_i$ be isomorphic for other choices.
However, when using a machine learning-based RE method, the structure of the graph $\mathcal{G}(q(o_i))$ may differ for each sentence $q(o_i)$ because the method has probabilistic behavior depending on the input.
Therefore, to satisfy the aforementioned condition, we create a \textbf{propositional graph template}
$\mathrm{pg}(\#)$ by (i) replacing the node string $o_i$ in original ${\cal G}(q(o_i))$ with a special symbol $\#$ and (ii) joining these $\mathcal{G}(q(\#))$ together:
\begin{equation*}
    \mathrm{pg}(\#)\coloneqq \bigcup_{o_i\in \mathcal{O}}{\cal G}(q(o_i))\{o_i\coloneq\#\},
\end{equation*}
where $\{o_i\coloneq\#\}$ means the substituion of $\#$ to $o_i$.
Hereafter, $\mathrm{pg}(o_i)$, which is obtained by substituting $o_i$ for the special symbol $\#$, is used as the propositional graph for each choice, instead of the original $\mathcal{G}(q(o_i))$.

\noindent{\bf Constructing Knowledge Graph}
Based on Wikipedia articles, we construct a knowledge graph for verifying the propositional graph. Here, let the set of nodes in $\mathrm{pg}(o_i)$ be $V_P$.
If a human were to verify the proposition $q(o_i)$, they would search for the keywords that are the subject of the sentence on Wikipedia and check if the content of the article matches the claim of the proposition.
Our method mimics this process by treating each node label in $\mathrm{pg}(o_i)$ as a topic.
Therefore, for each node $v\in V_P$, its label $\mathrm{label}(v)$ is searched in Wikipedia and the resulting article text $\mathrm{text}(v\, |\,  o_i)$ is converted into a relational graph.
\begin{align*}
    \mathrm{kg}(v\, |\, o_i) & \coloneqq \mathcal{G}(\mathrm{text}(v\,|\,o_i))
\end{align*}
$\mathrm{kg}(v\,|\,o_i)$ for each node $v\in V_P$ is then joined together to construct the knowledge graph
$\mathrm{kg}(o_i)$ representing the ground truth triplets:
\begin{equation*}
    \mathrm{kg}(o_i) \coloneqq \bigcup_{v\in V_P}\mathrm{kg}(v\,|\,o_i).
\end{equation*}

\noindent{\bf Resolving Node Label Ambiguity}
Because of the notational fluctuations in natural language texts, sometimes multiple nodes in the relational graphs represent the same entity in the real world.
In order to resolve such node label ambiguity, we apply the entity linking method for each node in the constructed graphs and replace original labels with entity titles if found.
By representing this process with the notation $\mathcal{L}$, the PG and KG for each choice $o_i$ are given as
\[
    \mathrm{PG}(o_i) = \mathcal{L}(\mathrm{pg}(o_i)), \mbox{ and }
    \mathrm{KG}(o_i) = \mathcal{L}(\mathrm{kg}(o_i)).
\]

\paragraph{Step 2.}
Let the node sets for $\mathrm{PG}(o_i)$ and $\mathrm{KG}(o_i)$ be $V_P$ and $V_K$.
In the following, we verify the correctness of the proposition $q(o_i)$ by comparing $\mathrm{PG}(o_i)$ and $\mathrm{KG}(o_i)$ created in Step 1. The verification is based on the idea that the more edges $(v_1, r, v_2)\in \mathrm{PG}(o_i)$ are included in $\mathrm{KG}(o_i)$, the more likely the original sentence $q(o_i)$ is correct.

However, since these graphs are created by RE, even nodes referring to the same entity may have variations in their labels at this time, depending on the original words in the sentences.
Therefore, we need to find the corresponding nodes between the subset of $V_P$ and $V_K$, namely $V_P' \coloneqq V_P\setminus V_K'$ and $V_K' \coloneqq V_K\setminus V_P$, respectively, excluding the nodes with perfectly matched labels.
Then, we compute the mapping $\varphi'\colon V_P'\rightarrow V_K'$ that maximizes the sum of sentence similarity between corresponding node labels:
\begin{equation*}
    \hat{\varphi}' \coloneqq \arg\max_{\varphi'} \sum_{v\in V_P'} \mathrm{sim}\left(\mathrm{label}(v),\mathrm{label}(\varphi'(v))\right).
\end{equation*}
Here, $\varphi'\colon V_P'\rightarrow V_K'$ represents a \textbf{bijection} of node correspondences, and $\mathrm{sim}(\cdot)$ represents the sentence similarity between two labels. This problem is a maximum matching problem in a bipartite graph consisting of two node sets $V_P'$ and $V_K'$. Using the solution $\hat{\varphi}'$, the desired correspondence between the entire node sets $\hat{\varphi}\colon V_P\rightarrow V_K$ is defined as
\begin{equation*}
    \hat{\varphi}(v) \coloneqq \begin{cases}
        \hat{\varphi}'(v) & \text{if } v\in V_P',        \\
        v                 & \text{if } v\in V_P\cap V_K.
    \end{cases}
\end{equation*}

\paragraph{Step 3.}
Finally, we count the edges $(v_1, r, v_2)\in \mathrm{PG}(o_i)$ that are also included in $\mathrm{KG}(o_i)$, and define \textbf{truthfulness (edge score)} of the original proposition $q(o_i)$ as the ratio of mutually included edges.

We define a projected version of $\mathrm{PG}(o_i)$, namely $\hat{\varphi}(\mathrm{PG}(o_i))$, by replacing the nodes $v_1,v_2\in V_P$ with the corresponding nodes $\hat{\varphi}(v_1),\hat{\varphi}(v_2)\in V_K$ as
\begin{align*}
    \hat{\varphi}(\mathrm{PG}(o_i)) & = \left\{
    \left(\hat{\varphi}(v_1), r, \hat{\varphi}(v_2)\right)
    \,\middle|\,
    \left(v_1, r, v_2\right) \in \mathrm{PG}(o_i)
    \right\}.
\end{align*}
Using this notation, the desired truthfulness $\mathcal{T}$ is expressed by the following equation:
\begin{align*}
    \mathcal{T}(o_i) \coloneqq \frac{\left|\hat{\varphi}(\mathrm{PG}(o_i))\cap \mathrm{KG}(o_i)\right|}{\left|\mathrm{PG}(o_i)\right|} \in [0,1]
\end{align*}
Here, $\left|G\right|$ represents the number of semantic triplets in a relational graph $G$.

\paragraph{Step 4.}
By applying the above three steps to $q(o_i)\ (i=1,2,3,4)$, we estimate the correct option
\begin{align*}
    \hat{o} \coloneqq \arg\max_{o_i\in \mathcal{O}} \mathcal{T}(o_i).
\end{align*}
If there are multiple $o_i$ that give the maximum value for $\mathcal{T}(o_i)$, we also take the average label similarity of the corresponding nodes $\mathcal{N}(o_i)$ into account in addition to $\mathcal{T}(o_i)$.
\begin{equation*}
    \mathcal{N}(o_i) \coloneqq \frac{1}{|V_P|}\sum_{v\in V_P} \mathrm{sim}(\mathrm{label}(v),\mathrm{label}(\hat{\varphi}(v))) \label{math-estimation-by-node}
\end{equation*}
We call this $\mathcal{N}(o_i)$ the \textbf{node score}.
If $\hat{o}$ cannot be narrowed down to one, the method randomly selects an option from the choices that have the highest edge and node scores.

\section{Experiments}\label{sec:experiments}
In this section, we conduct answering experiments on original MCQ datasets to demonstrate the effectiveness and the traceability of our method.
We also compare the performances of various RE methods over different lengths of question sentences to investigate the method's capability on each question category.

\subsection{Experimental Setup}
\paragraph{MCQ Datasets}
For the answering experiment, we created two variants of the MCQ dataset in the ``fill-in-the-blank" format, namely KR-200m and KR-200s, using GPT-4o \cite{openai_gpt-4o_2024}.
Both datasets consist of 10 categories shown in Figure \ref{fig:accuracy-rebel}, and each category has 20 questions.
The difference between KR-200m and KR-200s is the length of the question sentence $q(x)$.
KR-200m has 20.1 words in $q(x)$ on average, including blank position $x$ counted as one word, while KR-200s has only 7.5 words per sentence.

\paragraph{Relation Extraction Methods}
For the used RE methods, we consider REBEL \cite{huguet_cabot_rebel_2021}, mREBEL$_{400}$, mREBEL$_{32}$ \cite{huguet_cabot_redfm_2023}, and UniRel \cite{tang_unirel_2022} trained on the NYT dataset \cite{hutchison_modeling_2010}.

\paragraph{Entity Linking Method}
For the simplicity of the implementation, we construct an EL method using the title search of the Wikipedia API.
In this method, we search for the original node labels in Wikipedia and link them to the article's title, which the API returns as the most relevant.
If no page is found, we do not assign any page title to the node, leaving the original node label.

\paragraph{Semantic Similarity Measure}
For the semantic similarity measure, we use a sentence embedding model all-MiniLM-L6-v2 \footnote{\url{https://huggingface.co/sentence-transformers/all-MiniLM-L6-v2}}, which converts a sentence $s$ into a 384-dimensional embedding vector, and define the similarity $\mathrm{sim}(v_1, v_2)$ between nodes $v_1$ and $v_2$ as the cosine similarity of their embedding vectors.
all-MiniLM-L6-v2 is a model trained on a dataset of over one billion sentences, based on the lightweight language model MiniLM \cite{wang_minilm_2020}.

\paragraph{Wikipedia Articles}
All Wikipedia pages are fetched on July 9th, 2025, via MediaWiki API endpoint\footnote{\url{https://www.mediawiki.org/wiki/API:Main_page}}.
For creating KGs, we used the summary part of Wikipedia articles, which appears before any sections on the page.

\subsection{Results}
\paragraph{Overall Accuracy}
\begin{table}
    \centering
    \begin{tabular}{llrr}
        \toprule
                       & EL Setting & KR-200m       & KR-200s       \\
        \midrule
        REBEL          & w/ EL      & \textbf{53.5} & 48.0          \\
                       & -          & 42.2          & 42.0          \\
        mREBEL$_{400}$ & w/ EL      & 52.2          & \textbf{49.7} \\
                       & -          & 43.3          & 41.6          \\
        mREBEL$_{32}$  & w/ EL      & 49.0          & 43.6          \\
                       & -          & 43.3          & 37.7          \\
        UniRel         & w/ EL      & 28.6          & 26.9          \\
                       & -          & 29.0          & 27.3          \\
        \bottomrule
    \end{tabular}
    \caption{Overall results of answering experiments by various RE methods, with and without entity linking (EL) methods applied in graph creation. The best results across different datasets and settings are highlighted in \textbf{bold}. Accuracy (\%) is used as the evaluation metric.}
    \label{tab:overall-result}
\end{table}
The overall results of answering experiments across various RE methods and EL settings are shown in Table \ref{tab:overall-result}.
According to the table, REBEL and its variant mREBEL$_{400}$ achieve the highest accuracy on the two datasets, offering 53.5\% and 49.7\% accuracy on KR-200m and KR-200s, respectively.
Following the two methods, mREBEL$_{32}$ performs at slightly lower accuracy, while UniRel merely answers with as much accuracy as a random chance.
These results show that each model's number of relation types has an essential effect on accuracy.
However, despite the vast difference between the number of relation types that REBEL and mREBEL$_{400}$ can generate (220 and 400, respectively), both models perform almost equally on the two datasets, indicating that too large a number of relation types does not improve the result anymore.
In addition, the use of EL enhances the accuracy for three RE methods except UniRel, exhibiting the importance of node label disambiguation in the graph creation process.

\paragraph{Results by Category}
\begin{figure*}[tb]
    \centering
    \includegraphics[scale=0.48]{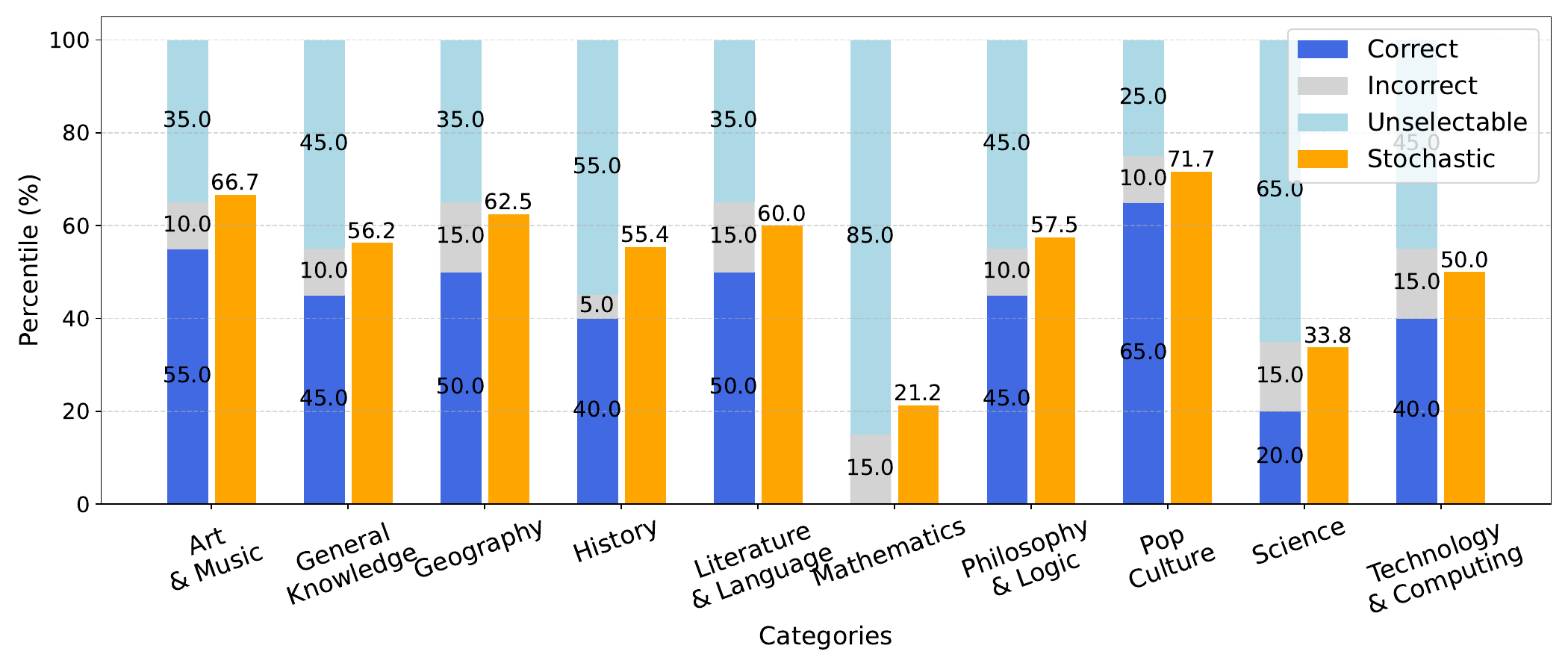}
    \caption{Accuracy per category for REBEL with entity linking on KR-200m. The columns with \textit{Correct}, \textit{Incorrect}, and \textit{Unselectable} show the results without random choice, determined solely by the edge and node scores. \textit{Stochastic} shows the accuracy with random selection for the unselectable questions.}
    \label{fig:accuracy-rebel}
\end{figure*}
Figure \ref{fig:accuracy-rebel} shows the accuracy per category for REBEL with EL enabled, which performed the best score on the KR-200m dataset.
Among the 10 categories, our method achieved the highest accuracy of 71.7\% on ``Pop Culture", followed by 7 categories with 50.0\% to 66.7\% accuracy.
In contrast, the method did not perform well in ``Mathematics" and ``Science".
These results indicate that our method achieves higher accuracy in categories where factual knowledge, represented as semantic triplets, is required to select the best option.
On the other hand, it does not perform well on questions that require abstract knowledge or logical inferences, such as mathematical derivations.

\subsection{Case Study of Traceability}
For a more intuitive understanding of the results, we illustrate PGs and KGs for an MCQ.
The question is from KR-200m, ``Art \& Music" category, and the question sentence $q(x)$ and the set of option words $\mathcal{O}$ are shown in Example \ref{ex:starry-night}.
\begin{example}[Starry Night]\label{ex:starry-night}
    Let $q(x)$ be \textit{``Vincent van Gogh, a Dutch post-impressionist painter, created several masterpieces, including the famous painting called $\{x\}$,"} where
    \begin{equation*}
        \mathcal{O} = \left\{\begin{aligned}
            o_1 & \colon\text{``Starry Night"},                                 \\
            o_2 & \colon\text{``The Persistence of Memory"},                    \\
            o_3 & \colon\text{``Guernica"},\quad o_4 \colon\text{``The Scream"}
        \end{aligned}\right\}.
    \end{equation*}
\end{example}
\begin{figure}[tb]
    \centering
    \begin{minipage}[b]{0.48\linewidth}
        \centering
        \includegraphics[scale=0.149]{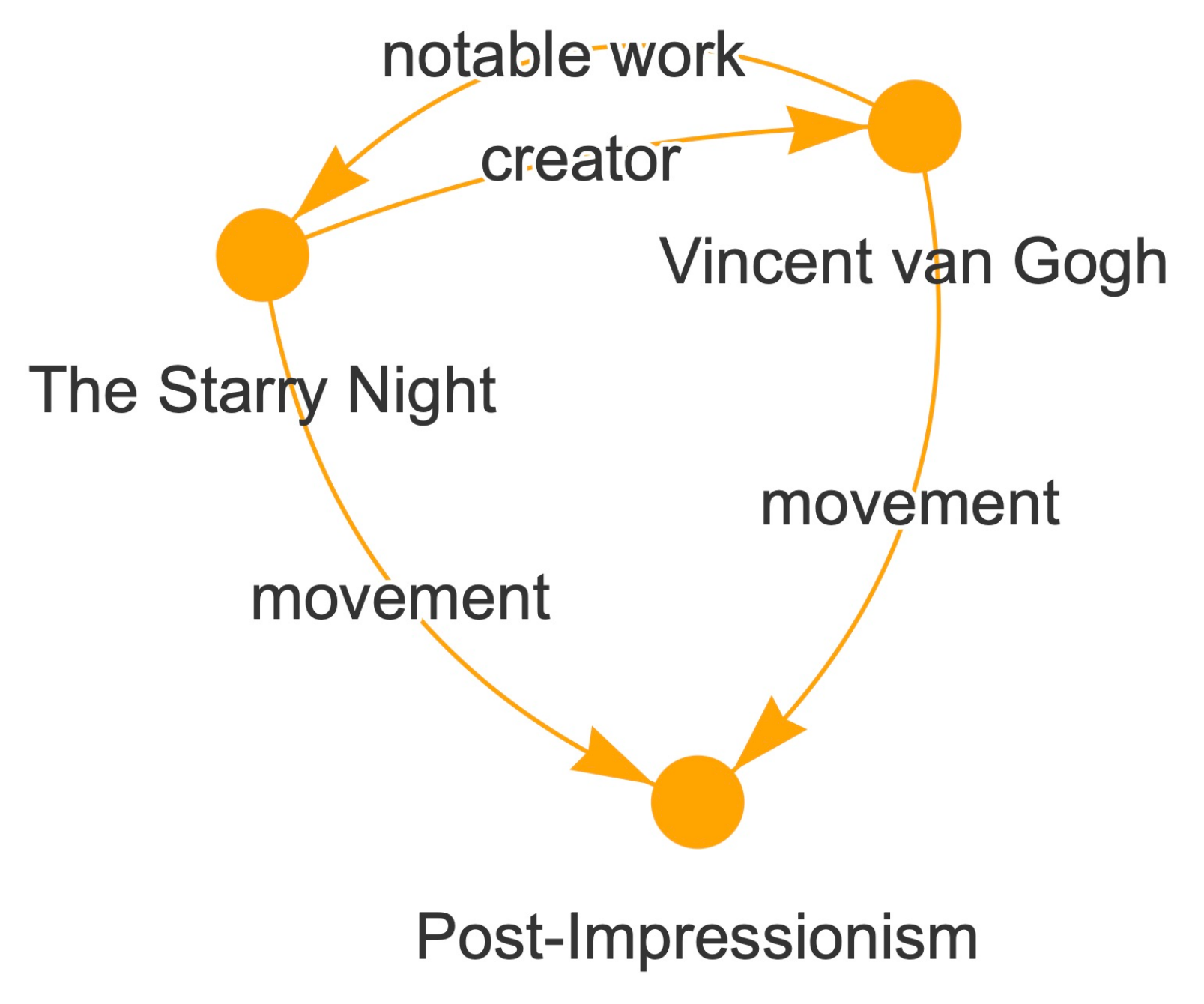}
    \end{minipage}
    \hspace{0.02\linewidth}
    \begin{minipage}[b]{0.48\linewidth}
        \centering
        \includegraphics[scale=0.15]{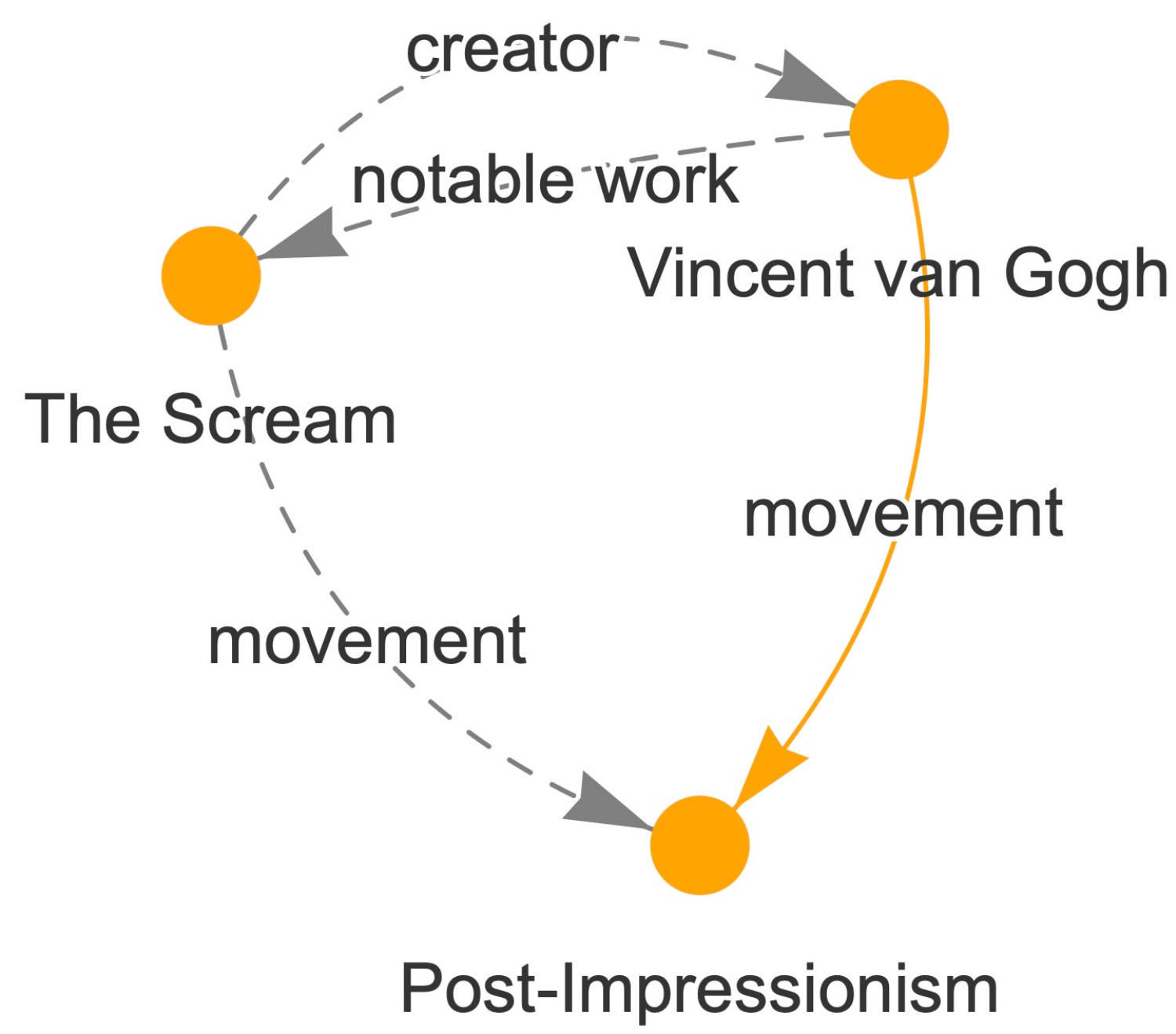}
    \end{minipage}
    \caption{PGs for the \textbf{correct} option $o_1\colon$``Starry Night" (left) and the \textbf{incorrect} option $o_4\colon$``The Scream" (right).}
    \label{fig:art1-pgs}
\end{figure}
\noindent
Figure \ref{fig:art1-pgs} shows the PGs for the correct option $o_1$ and the incorrect option $o_4$, respectively.
Edges shown as solid lines are the ones verified from KGs.
Comparing these two graphs, we can understand that the correct option $o_1\colon$``Starry Night" was selected because all relations in $\mathrm{PG}(o_1)$ are verified,
whereas only one edge is verified for $\mathrm{PG}(o_4)$.
Through such a comparison between the PGs for the chosen option and the one not selected, our method offers traceability on why a certain choice was selected or \textbf{not} selected.

\section{Related Works}\label{sec:ex-research}
In this section, we explain previous research on \textbf{multiple-choice question answering (MCQA)} and \textbf{fact verification}, which predicts if a given sentence is factually correct or not.

Most previous methods tackling MCQA naturally involve LLMs, because existing datasets, including MMLU \cite{hendryckstest2021} and BIG-bench \cite{srivastava2023beyond}, are constructed for evaluating the performance of LLMs.
On the other hand, a combination of KGs and language models has been proposed to enhance reasoning capabilities.
A method is proposed by \cite{zhang_joint_2023} that retrieves subgraphs from a knowledge graph database tailored for the question sentence $q$ and the set of options $\cal O$ and inputs them into a Graph Neural Network (GNN) to obtain answers.
This method is similar to our proposed method in that it creates KGs tailored to the problem setting.
However, it differs from ours because it targets question formats other than the ``fill-in-the-blank" and utilizes language models and reasoning mechanisms.
Additionally, since it uses GNNs, there is a problem of low explainability regarding the process until the output results are obtained.

\textbf{Fact verification} is a task that determines whether a natural language sentence is factually correct, also known as fact-checking \cite{guo_survey_2022}.
Although fact verification is a natural language task, language models lack clarity in the process leading to the answer, resulting in low explainability.
Therefore, methods using KGs with explicit link structures have been proposed. Yuan and Vlachos \shortcite{yuan_zero-shot_2024} proposed a method that extracts semantic triplets $(\mathrm{subject}, \mathrm{relation}, \mathrm{object})$ from sentences of unknown veracity and verifies them against a knowledge graph database.
This study aligns with ours in converting sentences to be verified into a list of triplets.
On the other hand, the method for preparing the ground truth knowledge graph differs, and they utilized additional reasoning modules.\par

Tackling MCQs on the cloze test format is positioned as a special case of MCQA. However, in MCQA, many previous studies use LLMs without the interest of interpretability, and methods pursuing traceability are in the minority.
Therefore, in this study, we proposed an explainable MCQA method using KG-based fact verification.

\section{Conclusion}\label{sec-conclusion}
In this paper, we presented a combination of the RE method and KG matching for answering MCQs in the ``fill-in-the-blank" format.
In our framework, the sentence, which is formed by substituting a choice word into the blank position of the question, is regarded as a proposition.
We proposed a fact verification method by converting the sentence into a relational graph and verifying it against a factually correct KG.
We conducted the answering experiments using various RE methods in two original MCQ datasets with different question lengths.
Our method was able to answer up to 70\% of the questions, depending on the category, while keeping the traceability of the chosen answer by visualizing verified edges in propositional graphs and the knowledge graph triplets used for verification.
The result comparison over various RE methods reveals the tendency for a greater number of relation types in the RE method to cause higher accuracy.
In addition, the ablation study demonstrates the importance of the entity linking method in the graph creation process.
Future research could improve the accuracy of article search in the KG creation process in order to prevent the mislinking of Wikipedia pages and introduce a more flexible verification mechanism to implement reasoning capability in the method.

\section*{Acknowledgments}
This work was supported by JSPS KAKENHI Grant Number JP25H01112.
The authors would also like to thank Assistant Professor Nozomi Akashi for his valuable comments and feedback on the research.

\section*{Code and Data Availability}
The source code and dataset used in this study are publicly available at: \url{https://github.com/N-Shimoda/KG-MCQA}.

\bibliographystyle{kr}
\bibliography{kg-mcqa}
\end{document}